\documentclass{article}

    \PassOptionsToPackage{numbers, compress}{natbib}

      \usepackage[final]{ai4earth_final}

\usepackage{booktabs}       
\usepackage{graphicx}

\title{Developing High Quality Training Samples for Deep Learning Based Local Climate Zone Classification in Korea}

\author{
  Minho Kim{\textsuperscript{*}}\\

  Seoul National University\\
  mhk93@snu.ac.kr  \\
  \And
  Doyoung Jeong{\textsuperscript{*}}\\
  Seoul National University\\
  tmits37@snu.ac.kr  \\
  \And
  Hyoungwoo Choi{\textsuperscript{*}}\\
  Seoul National University\\
  chhw95@snu.ac.kr  \\
  \And
  Yongil Kim{\thanks{Department of Civil and Environmental Engineering}}\\
  Seoul National University\\
  yik@snu.ac.kr \\
}

\begin{document}

\maketitle

\begin{abstract}
Two out of three people will be living in urban areas by 2050, as projected by the United Nations, emphasizing the need for sustainable urban development and monitoring. Common urban footprint data provide high-resolution city extents but lack essential information on the distribution, pattern, and characteristics. The Local Climate Zone (LCZ) offers an efficient and standardized framework that can delineate the internal structure and characteristics of urban areas. Global-scale LCZ mapping has been explored, but are limited by low accuracy, variable labeling quality, or domain adaptation challenges. Instead, this study developed a custom LCZ data to map key Korean cities using a multi-scale convolutional neural network. Results demonstrated that using a novel, custom LCZ data with deep learning can generate more accurate LCZ map results compared to conventional community-based LCZ mapping with machine learning as well as transfer learning of the global So2Sat dataset.

\end{abstract}

\section{Introduction}

The Local Climate Zone (LCZ) classification framework was introduced as a climate-based classification of urban-rural areas for temperature studies \cite{stewart2012local}. The LCZ scheme (see Appendix, Figure 5) is composed of 17 urban and natural classes based on regions of uniform surface cover, structure, material, and anthropogenic activity \cite{demuzere2020combining}. In general, LCZs are culturally-neutral and are characterized by screen-height temperature apparent over dry surfaces on calm, clear nights in areas of simple relief \cite{stewart2012local}. Based on this concept, the integration of remote sensing data and data-driven methods have enabled global LCZ classification at 100 m spatial resolution \cite{bechtel2015mapping}.

\subsection{Global Trends in LCZ Classification}

The World Urban Database and Access Portal Tools (WUDAPT) is a community-based project which provided an automated LCZ classification protocol to produce LCZ maps of global cities using the random forest (RF) classifier. \cite{bechtel2015mapping,bechtel2019generating}. Recent studies have extended LCZ mapping to national and continental scales, such as the United States \cite{demuzere2020combining}, Europe \cite{demuzere2019mapping}, Hong Kong \cite{wang2018mapping}, and major cities in China \cite{ren2019assessment}. 
Another approach in literature was improving the classification accuracy when employing deep learning over conventional machine learning algorithms \cite{yoo2019comparison,qiu2018feature,liu2020local,rosentreter2020towards}. Furthermore, Zhu et al. (2019) introduced the large-scale So2Sat dataset which is comprised of Sentinel-1 and 2 image patches with corresponding LCZ labels for 42 global cities. The dataset has been tested and benchmarked using sophisticated deep learning models, demonstrating superior classification accuracy results over conventional methods \cite{zhu2019so2sat, qiu2020multi}. However, the WUDPAT protocol is limited to low classification accuracy and poor generalization ability \cite{yoo2019comparison, zhu2019so2sat, bechtel2017quality}, while the So2Sat dataset requires advanced domain adaptation techniques to fully harness its potential.

\subsection{Deep Learning Based LCZ Mapping for Korea}

This study aims to develop LCZ mapping for the Korean peninsula by developing LCZ training samples of major Korean cities and classifying using a multi-scale convolutional neural network (MSCNN). The major contributions of this study can be summarized by the following points:

    \textbf{(1) High-quality LCZ training samples for Korea}: Adopted sampling method \cite{bechtel2015mapping, zhu2019so2sat} for scene classification via deep learning \cite{yoo2019comparison, liu2020local, rosentreter2020towards} to produce fundamental training samples for Korea.

    \textbf{(2) Assessment of classification accuracy}: Evaluated classification accuracy using RF and MSCNN models with the custom LCZ training samples and transfer learning of the state-of-the-art So2Sat dataset \cite{zhu2019so2sat}, thereby investigating the effectiveness of custom and global LCZ datasets

\section{Data and Methods}

\subsection{Input Data}

The sampling phase requires a basemap, natural cover, and built-up cover layers. The specifications of each data type is displayed in Table 1. Sentinel-2 images were used as the main basemap and were acquired using the Copernicus Open Hub. Google Earth Engine cloud-free mosaics can also be used \cite{schmitt2019aggregating}. The Normalized Difference Vegetation Index (NDVI) is applied as the natural cover layer and the master building information (MBI) dataset, provided by the Electronic Architectural Administration Information System (EAIS), is used as the built-up cover. The MBI dataset includes building information of each major city in Korea such as building height, number of floors, area, and location. These input layers can be combined to help label sampling (see Appendix, Figure 6).

\begin{table}[ht]
  \caption{List of input data specifications}
  \label{sample-table2}
  \centering
  \begin{tabular}{ccc}
    \toprule
    Data &  Main Function & Source \\
    \midrule
    Sentinel-2 & Basemap and input (Bands 1 - 10) & Copernicus Open Hub \\
    NDVI & Natural cover (LCZ A - LCZ G) & Sentinel-2 Bands 3 and 4 \\
    MBI Building height & Built-up cover (LCZ 1 - LCZ 6) & EAIS \\
    \bottomrule
  \end{tabular}
\end{table}

\subsection{Sampling Workflow}

This study adopted a scene classification approach to create 32 by 32 pixel patches based on Liu et al. (2020) and to match the So2Sat's patch size. The entire labeling workflow (see Appendix, Figure 7) is based on the workflow in Zhu et al. (2019). In this study, the LCZ labels are sampled as points which act as the patch's centroid \cite{ liu2020local, rosentreter2020towards, zhu2019so2sat}. Larger image patches tend to improve classification accuracy due to local information \cite{yoo2019comparison, liu2020local, rosentreter2020towards}. Second, the LCZ labels are guided using MBI and reviewed with Google Earth and Google Street View to ensure correctness and completeness \cite{zhu2019so2sat}. Finally, once a sufficient volume of training samples is accumulated, data augmentation is processed to modify class weights to alleviate the effect of imbalanced classes.

\subsection{Assessment of Classification Accuracy}

First, the RF is used to represent the WUDAPT method. Second, MSCNN is built with a multi-scale layer which uses three convolution layers with different kernel sizes (3, 5, 7 pixels), followed by five convolution layer blocks in order of convolution (CONV), batch normalization, ReLU activation function, and max pooling layers. A dropout out of 25\% was added after the first fully-connected (FC) layer. MSCNN used the adam optimizer \cite{kingma2014adam}, batch size of 96, learning rate of 0.002, decay factor of 0.004, and early stop of 15. Third, the So2Sat dataset uses the benchmark ResNet for training \cite{zhu2019so2sat}. The original layers are frozen, while two fully-connected dense layers are added to the end of the model for transfer learning \cite{zhao2017transfer}. 

\begin{figure}[ht]
  \centering
  \includegraphics[width=1\linewidth]{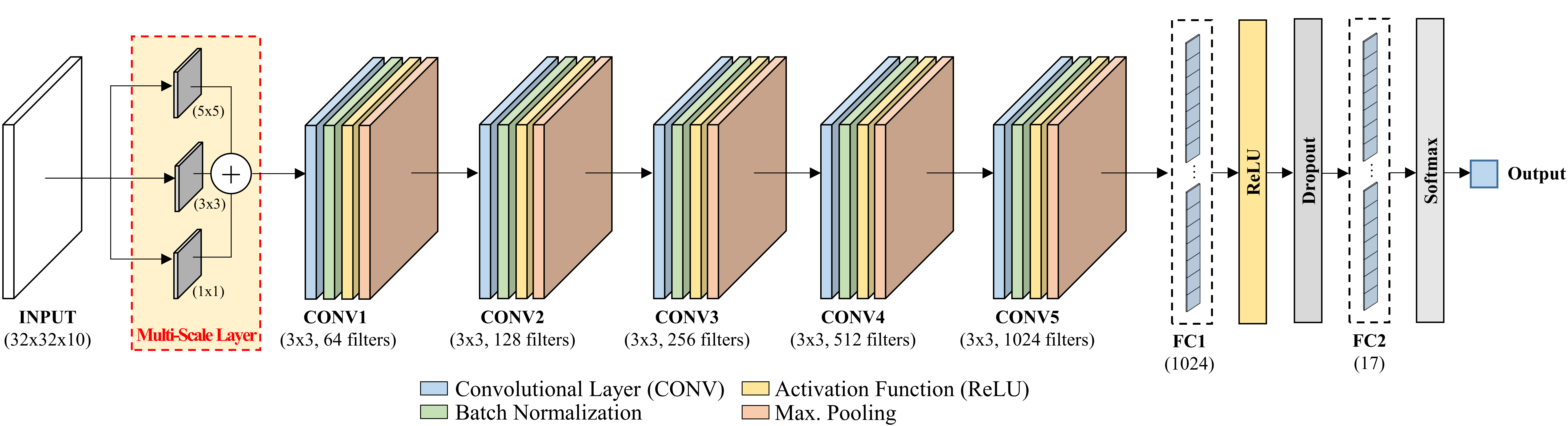}
  \caption{Model architecture of MSCNN showing the multi-scale layer and convolutional layers.}
\end{figure}

\section{Results and Discussion}

\subsection{Contribution 1: High-quality LCZ training samples for Korea}

The distribution of training samples for major Korean cities are provided in Figure 2 with respect to each LCZ class. The different class proportions offer a glimpse of each city's urban form and structure. This list is not exhaustive and we plan to continue increasing data volume and labeling quality. These training samples can then be used as the fundamental training sample dataset to generate a nation-level LCZ map of Korea.

\begin{figure}[ht]
  \includegraphics[width=0.8\linewidth]{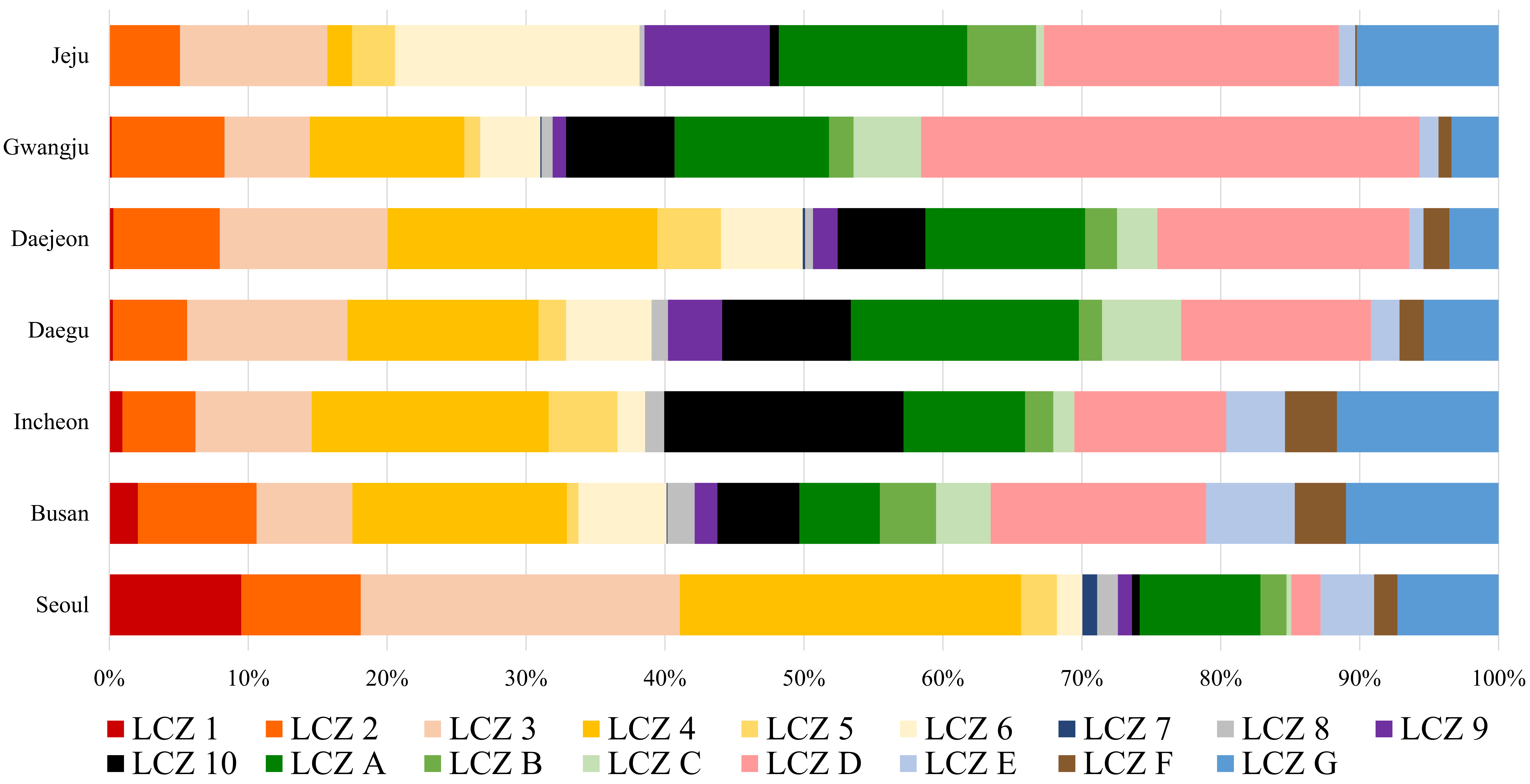}
  \centering
  \caption{Current distribution of labelled LCZ training samples of key cities.}
\end{figure}

\subsection{Contribution 2: Classification Accuracy}
LCZ classification was conducted for Seoul to produce 100 m resolution LCZ maps using the three models mentioned in Section 2.3. As organized in Table 2, MSCNN using the custom LCZ data outperformed all other methods with a superior overall accuracy of 83.88\%. Compared to RF when using the custom data (74.27\%), MSCNN demonstrates that deep learning is significant to improve accuracy, while the low accuracy of So2Sat (54.42\%) implies that using MSCNN with the custom data is more effective than transfer learning a global dataset, in spite of the large-scale data volume.

\begin{table}[ht]
  \caption{LCZ classification accuracy}
  \label{sample-table1}
  \centering
  \begin{tabular}{cccc}
    \toprule
    Model     & Overall     & Kappa \\
    \midrule
    So2Sat & 54.42\%  & 48.60\%    \\
    RF & 74.27\%   & 41.00\%  \\
    \textbf{MSCNN} & \textbf{83.88\%}     & \textbf{64.99\%}   \\

    \bottomrule
  \end{tabular}
\end{table}

\begin{figure}[ht]
  \includegraphics[width=1\linewidth]{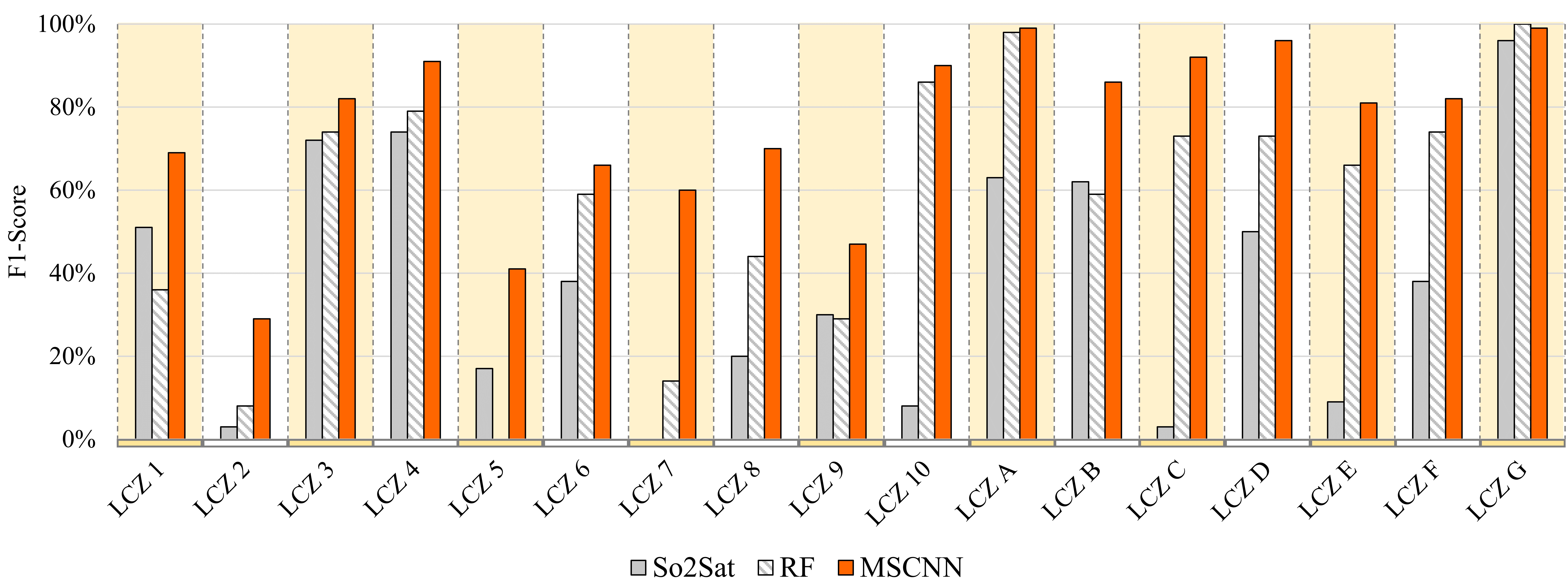}
  \centering
  \caption{Class-wise comparison of F1-score results.}
\end{figure}

The F1-score results shown in Figure 3 indicate that MSCNN classified the highly heterogeneous features more effectively compared to RF and the So2Sat method. For instance, the built-up LCZ classes, such as LCZ 1 or LCZ 4 were more defined using MSCNN, especially in densely compact urban areas and even suburban and rural communities. The So2Sat method was able to classify built-up areas like LCZ 1, but transfer learning of the global dataset introduced confusion of natural cover in densely vegetated areas. This result can be attributed to the difference in crop fields and covers between global cities and Korea, which is particularly prevalent by seasonal variations. All of the models struggled to classify similar LCZ classes (below 50\% F1-score), such as LCZ 2 and 3 as well as for LCZ 5 and 6 \cite{qiu2018feature}. Future studies can integrate DSMs or building height information from GIS layers such as OpenStreetMap and the EAIS dataset.

\begin{figure}[ht]
  \includegraphics[width=1\linewidth]{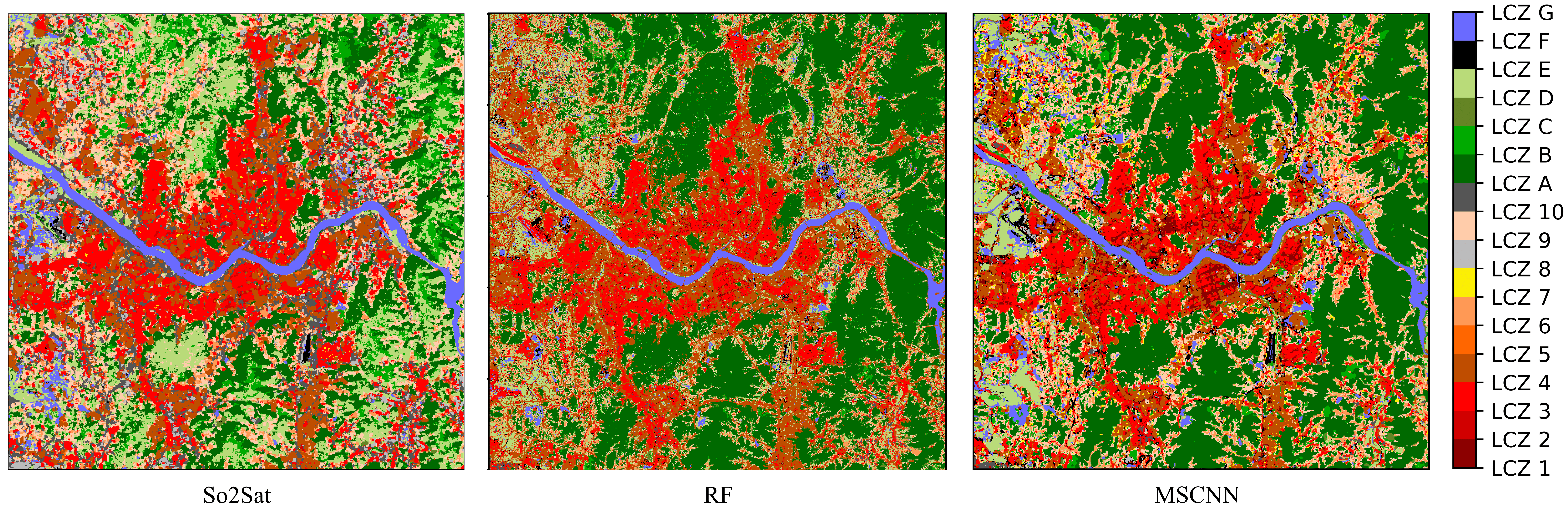}
  \centering
  \caption{Comparison of LCZ maps of Seoul for May 3, 2017 for So2Sat, RF, and MSCNN.}
\end{figure}

\section{Outlook and Future Work}

Deep learning based LCZ classification methods are required to effectively monitor urban areas and to produce high quality LCZ maps that delineate the heterogeneous urban features effectively. Based on the progress of generating global-scale LCZ maps and datasets in recent studies, this study developed fundamental, custom data for LCZ mapping in Korea. The study's results using MSCNN confirm that deep learning outperforms the shallow RF classifier used by WUDAPT, and that creating a novel, yet optimized labelled training sample data can be more efficient compared to transfer learning the large-scale So2Sat dataset. Future works will focus on increasing data volume and enhancing label quality, while exploring ways to integrate global datasets such as So2Sat more effectively. City-by-city comparisons of LCZ classification will also be conducted to investigate the generalization ability of the models and training sample data.

\bibliographystyle{unsrt}
\bibliography{2020_NIPS_AI4EARTH.bib}

\section{Appendix}
The following figures are included to help users understand and replicate our sampling and classification procedure. Despite the difficulty of manually labeling training samples, we believe that optimizing and enhancing the input data is essential to ensure high quality LCZ maps.

\begin{figure}[ht]
  \centering
  \includegraphics[width=1\linewidth]{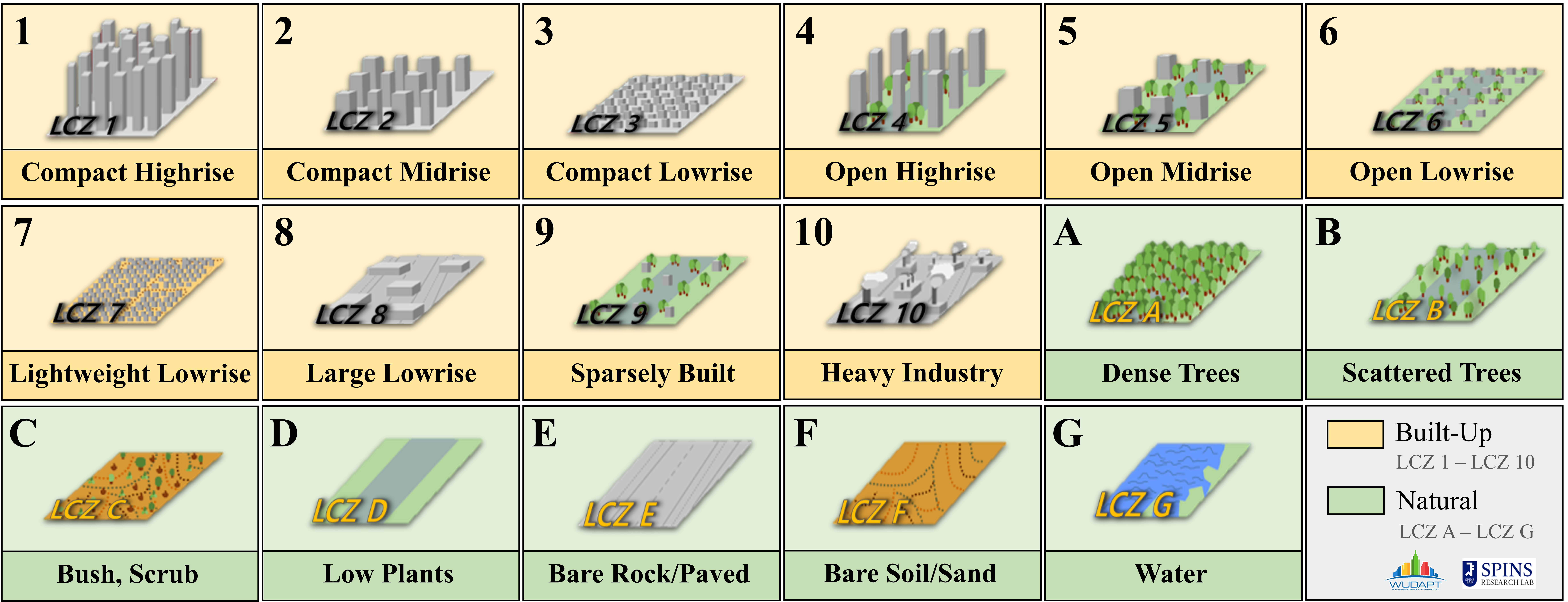}
  \caption{Overview of the LCZ framework with a description of each class. For more detailed explanations, readers are directed to more comprehensive studies \cite{stewart2012local, bechtel2015mapping}.}
\end{figure}

\begin{figure}[ht]
  \centering
  \includegraphics[width=1\linewidth]{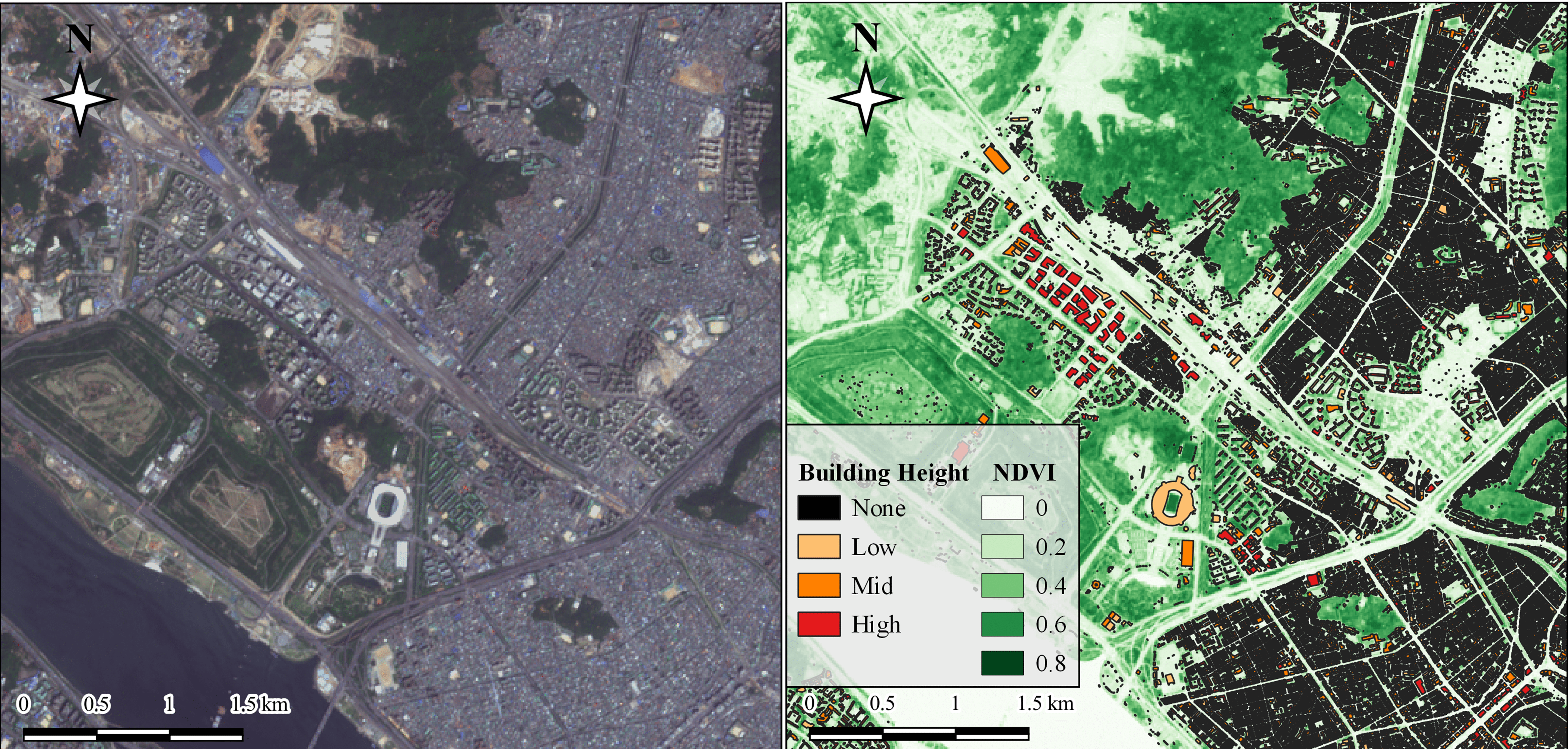}
  \caption{Example of a stacked layer analysis for LCZ sampling.}
\end{figure}

\begin{figure}[ht]
  \centering
  \includegraphics[width=1\linewidth]{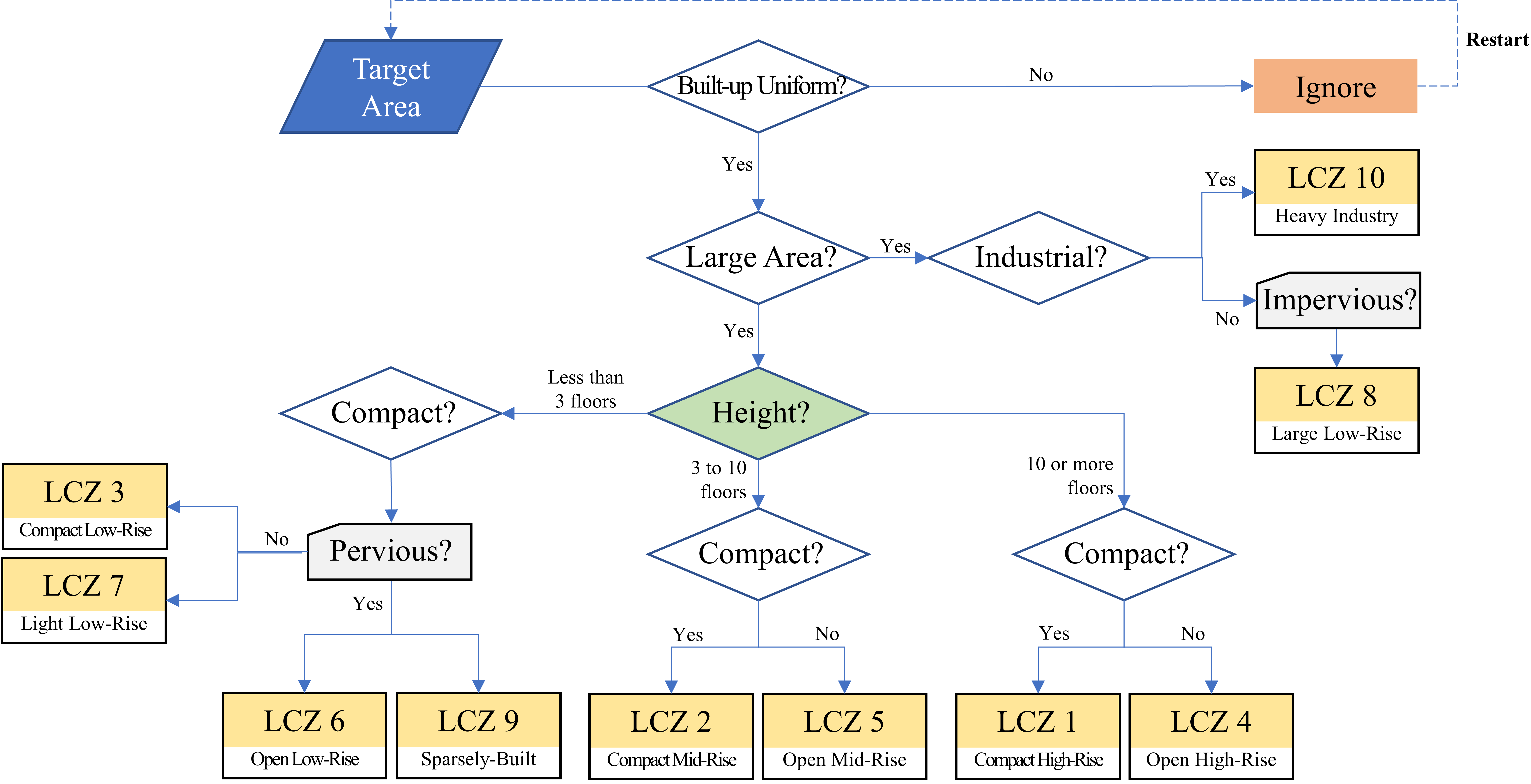}
  \caption{Detailed decision rule workflow for built-up LCZ class labeling adapted from Zhu et al. (2019). The main difference is the addition of the EAIS database to help differentiate building height as well as the consideration of pervious and impervious covers when classifying low-rise and industry-based LCZ classes.}
\end{figure}

\end{document}